# Deep GrabCut for Object Selection

Ning Xu[1]
ningxu2@illinois.edu

Brian Price[2]
bprice@adobe.com

Scott Cohen[2]
scohen@adobe.com

Jimei Yang[2]
jmyang@adobe.com

Thomas Huang[1]
huang@ifp.uiuc.edu

[1] University of Illinois at Urbana-Champaign
1401 W Green St, Urbana, IL 61801

[2] Adobe Research
345 Park Avenue, San Jose, CA 95110

## Abstract

Most previous bounding-box-based segmentation methods assume the bounding box tightly covers the object of interest. However it is common that a rectangle input could be too large or too small. In this paper, we propose a novel segmentation approach that uses a rectangle as a soft constraint by transforming it into an Euclidean distance map. A convolutional encoder-decoder network is trained end-to-end by concatenating images with these distance maps as inputs and predicting the object masks as outputs. Our approach gets accurate segmentation results given sloppy rectangles while being general for both interactive segmentation and instance segmentation. We show our network extends to curve-based input without retraining. We further apply our network to instance-level semantic segmentation and resolve any overlap using a conditional random field. Experiments on benchmark datasets demonstrate the effectiveness of the proposed approaches.

# 1 Introduction

Rectangles are often used as input in computer vision tasks. For example, Rother *et al*. [19] introduced using a rectangle around an object as input for interactive segmentation. In instance segmentation, many methods [3, 4, 5, 8, 13, 14] follow the detect-and-segment pipeline where several detection boxes are first obtained by some automatic detection algorithms, then segmentation is performed on each of the boxes independently. However, an important question raises: are rectangles leveraged in an effective way by current methods?

Fig. 1 shows examples of various rectangle cases and the segmentation results obtained by different algorithms. The results of the first two examples are obtained by interactive segmentation methods [19] and [12]. Most previous methods in this field assume the rectangle is a bounding box. Pixels are then sampled inside/outside the bounding box to estimate the foreground/background distributions. However, this strong assumption cannot handle cases such as Fig. 1(a) where the box is inside the object boundaries and Fig. 1(b) where the box does not leave enough background pixels.

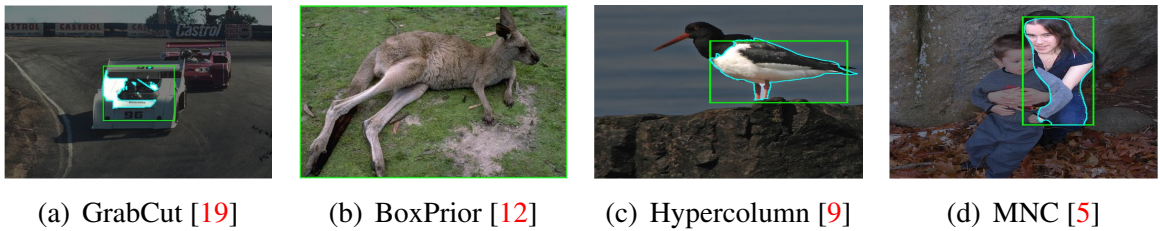

(a) GrabCut [19]  (b) BoxPrior [12]  (c) Hypercolumn [9]  (d) MNC [5]

Figure 1: Rectangles (in green) and the results obtained by different algorithms (in cyan). Note that the box in Fig. 1(b) contains the whole image. See Fig. 5 and 7 for our results.

The results of the last two examples are obtained by instance segmentation methods [9] and [5]. Most recent instance segmentation methods are based on deep neural networks. They all use detection rectangles as a hard constraint and either crop that region to use as the network input or pool network features inside the box. However, since the detection boxes are predicted by algorithms, many of them do not contain the whole object like in Fig. 1(c). Even if a detection covers most of an object like in Fig. 1(d), only using the information inside the rectangle will lose contextual information which is useful for segmentation.

The requirement to place fairly tight bounding boxes is too restrictive for user interaction and detection algorithms. In this paper, we propose a novel way to use rectangles for selection that produces more accurate results given a tight bounding box while also giving similar results for loosely-placed rectangles. Two key insights are that 1) a rectangle should suggests the object is nearby, and 2) global context information is important for correct segmentation. Inspired by [25], we transform the rectangle into a Euclidean distance map. The distance map is concatenated with the RGB image as input for a convolutional encoder-decoder network (CEDN) [1, 17]. Our model is trained with loose rectangles to allow for robust segmentation given sloppy input. The final prediction is the segmentation mask of the object. Transforming rectangles into distance maps satisfies our key insights. It encodes the distance from every pixel to the rectangle while also keeping all the image content. As a result, our model is robust to rectangle placement while previous methods are not. Moreover, our model is trained to segment general objects, and thus generates much more accurate results.

As an application, we apply our method to automatic multi-object segmentation by computing a selection from detection results. As multiple detections may overlap the same object, we apply a dense conditional random field (CRF) [11] to convert individual segments into an instance-level semantic labeling of the image. We also apply our model (trained on rectangles) to hand-drawn curves and show that our approach generalizes well to this input while allowing more flexibility in the input.

## 2 Related works

Computing a segmentation given a bounding box was introduced by Rother et al. [19]. It computes a selection by iterating between computing foreground/background color models given a segmentation and computing a segmentation given foreground/background color models. Improvements to this method include optimizing the segmentation and color models in one step [22] and adding a spatial extent prior [12]. However, these models all rely on basic color and edge information and do not use higher-order knowledge like the structure and shape of objects.

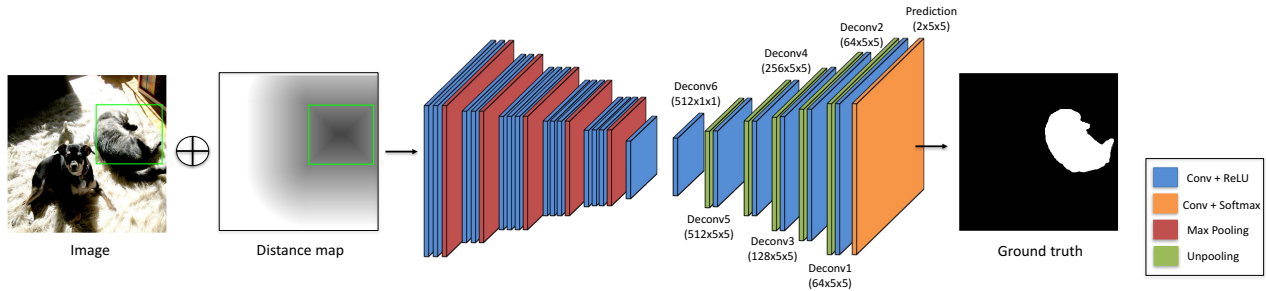

Figure 2: The framework of our segmentation model. The rectangle is indicated in green in the "Image" and "Distance map". The symbol ⊕ denotes the concatenation operation.

Detection-based instance segmentation methods also provide ways of computing a segmentation given a rectangle. The works of [5, 13, 14] are most relevant to ours as they use neural nets. Li et al. [13] trains a network that takes a segmentation heatmap as well as an image and produces a new heatmap. At test time, a heatmap is generated using [9] and then several iterations of the network are performed to get a solution. Dai et al. [5] use a network consisting of a region-of-interest pooling and two fully-connected layers to compute a mask given a bounding box. Liang et al. [14] train an instance-level segmentation neural network that takes an initial object proposal and a number of CNN feature maps and applies them iteratively to refine the segmentation. These methods each compute their segmentation within the bounding box and are thus sensitive to its placement. In contrast, we allow loosely-placed rectangles and generate a segmentation without need of a preliminary segmentation or multiple iterations of processing. Additionally, this line of work falls short of specifying exactly which pixels in an image belong to which object as they can return multiple bounding boxes for a single object, often with different class labels, and can also return overlapping segmentation for adjacent objects. We resolve this using a conditional random field to provide an instance-level semantic segmentation without overlap between objects.

## 3 Segmentation from rectangles

We propose a method of computing a segmentation from a rectangle, which is useful for both interactive segmentation and instance segmentation. In our approach, a rectangle is first transformed into a Euclidean distance map with the same size as the image input. Then the distance map is concatenated with the image along the channel dimension to construct an input pair to a CEDN model [1, 17]. The final prediction is the object mask. The framework of our approach is illustrated in Fig. 2. Our approach is inspired by [25] but different from theirs in mainly 4 aspects. First, [25] is not applicable for rectangle inputs. Second, they also do not address the sloppy input problem. Third, their method is a two-step process which requires post-processing by graph cut while our model is trained end-to-end. Last, we extend our segmentation results to automatic multi-object segmentation while there is no clear way to extend [25] to this problem.

### 3.1 Rectangle sampling

Since our segmentation model is trained on many (image, rectangle) pairs, our algorithm samples several rectangles for each instance by randomly jittering the ground truth bounding box. Specifically, for each instance, let us define the ground truth tight bounding box $\mathbf{B}^0$ as $[x^0_{\min}, y^0_{\min}, x^0_{\max}, y^0_{\max}]$ where each element represents the minimum/maximum $x/y$ coordinate. Our algorithm randomly samples $N_{\text{train}}$ rectangles. The four coordinates of rectangle $\mathbf{B}^i, i \in \{1, ..., N_{\text{train}}\}$ are generated by

$$
\begin{aligned}
x^i_{\min/\max} &= x^0_{\min/\max} + v \cdot g^i_j \cdot (x^0_{\max} - x^0_{\min}), \\
y^i_{\min/\max} &= y^0_{\min/\max} + v \cdot g^i_j \cdot (y^0_{\max} - y^0_{\min}),
\end{aligned}
\quad (1)
$$

where $g^i_j \sim \mathcal{N}(0, 1), j \in \{1, 2, 3, 4\}$ are standard Gaussian random variables. $v$ is a hyperparameter that controls the degree of rectangle variation. With this sampling strategy, a small training dataset can be augmented while keeping the model free from overfitting.

## 3.2 Rectangle transformation

Our algorithm then transforms each of the sampled rectangles into a distance map. In particular, given a rectangle $\mathbf{B}$ in an image $\mathcal{I}$, let us define the pixels on the edge of $\mathbf{B}$ as a set $\mathcal{S}_e = \{\mathbf{p}_i \mid \mathbf{p}_i \text{ is on the edge of } \mathbf{B}\}$ where $\mathbf{p}_i$ represent the location of pixel $i$. Similarly, let us define the pixels inside $\mathbf{B}$ as a set $\mathcal{S}_i$ and the pixels outside $\mathbf{B}$ as a set $\mathcal{S}_o$. Then our algorithm creates a 2-D distance map $\mathbf{D}$ which has the same width and height as the image $\mathcal{I}$. We compute $\mathbf{D}$ at location $\mathbf{p}_i$ as:

$$
\mathbf{D}(\mathbf{p}_i) = \begin{cases} 128 - \min_{\forall \mathbf{p}_j \in \mathcal{S}_e} |\mathbf{p}_i - \mathbf{p}_j|, & \text{if } \mathbf{p}_i \in \mathcal{S}_i, \\ 128, & \text{if } \mathbf{p}_i \in \mathcal{S}_e, \\ 128 + \min_{\forall \mathbf{p}_j \in \mathcal{S}_e} |\mathbf{p}_i - \mathbf{p}_j|, & \text{if } \mathbf{p}_i \in \mathcal{S}_o, \end{cases} \quad (2)
$$

where $|\cdot|$ denotes the Euclidean distance. We use signed distance transformation to better capture the context information of rectangle inputs while the unsigned transformation in [25] gives inferior results. For the efficiency of data storage, we truncate the values of $\mathbf{D}$ between 0 to 255. Finally we concatenate the RGB channels of $\mathcal{I}$ and the distance map $\mathbf{D}$ to construct a training pair.

## 3.3 CEDN model training

Our segmentation model inputs the concatenated training pairs and predicts binary instance masks. The structure is a CEDN model including an encoder and decoder. The encoder network is composed of several convolutional and max-pooling layers which abstract the input data to small feature maps while the decoder network has several convolutional and unpooling layers which reconstruct the image details from coarse to fine. We use the first 14 layers of VGG-16 [21] to initialize the encoder network. Since our input has 4 channels, the convolutional filters at the first layer have one extra channel compared to VGG-16 which are initialized with zeros. For the decoder network, we use a more concise network structure than the encoder network to reduce redundant parameters and speed up training. The decoder structure is shown in Fig. 2 in more details. All the parameters of the decoder network are initialized with Xavier [7]. To address overfitting, Dropout is used after each convolutional layer of the decoder network. In addition, we resample all the training data randomly in the beginning of each training epoch.

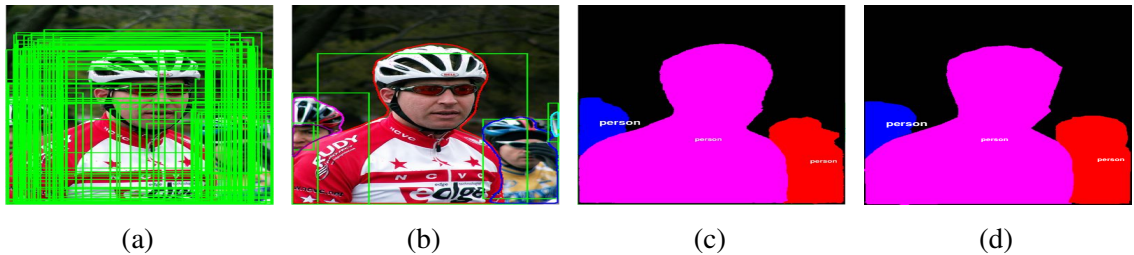

(a) (b) (c) (d)

Figure 3: The pipeline of our instance-level pixel labeling. (a) Image with all detection boxes. (b) The segmentation per detection after NMS. Object contours with different colors are outlined. (c) The final pixel-wise segmentation after CRF. (d) Ground truth.

## 4 Instance-level semantic segmentation

We extend our segmentation method to convert detection boxes into a instance-level semantic segmentation. In doing so, we not only compute a segmentation for each detection box independently but also resolve any overlap between segments to generate a pixel labeling.

Our approach to convert detection results into a pixel-labeling is illustrated in Fig. 3. Given a test image, our method takes as input a set of detections and accompanying class labels and scores. We first process the detections using non-max suppression ignoring the class labels. We use our rectangle segmentation on each remaining detection box $i$ independently to generate a foreground probability map $\mathbf{P}_i^f$ and a background probability map $\mathbf{P}_i^b$.

Our algorithm weights each detection box $i$ by using its largest categorical confidence score $s_i$ to update $\mathbf{P}_i^f$ and $\mathbf{P}_i^b$ such as $\mathbf{P}_i^{f\text{new}} = s_i \cdot \mathbf{P}_i^f$ and $\mathbf{P}_i^{b\text{new}} = 1 - \mathbf{P}_i^{f\text{new}}$. Since each $\mathbf{P}_i^{b\text{new}}$ represents the background probabilities given a particular rectangle $i$, the background probability map $\mathbf{P}^{b\text{new}}$ given all the rectangles can be computed as $\mathbf{P}^{b\text{new}} = \prod_{i=1}^N \mathbf{P}_i^{b\text{new}}$ where $N$ is the total number of detections. Finally we do normalization among $\mathbf{P} = [\mathbf{P}^{b\text{new}}, \mathbf{P}_1^{f\text{new}}, ..., \mathbf{P}_N^{f\text{new}}]$.

To assign an unique instance label to every pixel in the image, our algorithm solves the problem by leveraging the fully connected Conditional Random Field (CRF) model [11]. The objective function is formulated as $\min \sum_k \varphi_u(l_k) + \sum_{k<j} \varphi_p(l_k, l_j)$, where $l_k \in [0, N]$ is the label assignment for pixel $k$. We define the unary potential as $\varphi_u(l_k) = -\log(\mathbf{P}(l_k))$. The pairwise potential commonly penalizes the label disagreement between nearby pixels with similar colors. We adopt the same formulation as used in [2]. The final solution of the objective function gives an pixel-level instance labeling as well as semantic labeling.

## 5 Experiments

### 5.1 Implementation

Our models are trained on the PASCAL VOC 2012 dataset [6] and MS COCO dataset [15]. For PASCAL dataset, we adopt the same training and testing splits as used in [8, 9]. For MS COCO dataset, we use the 2014 *train-80k* dataset for training. In each training iteration, we randomly sample 4 bounding boxes per instance and make them a mini-batch. The sampling parameter $v$ is 0.15. The training inputs are resized to $320\times 320$. We use Adam [10] for optimization. The learning rate is $10^{-5}$. The threshold of our non-max suppression is 0.5.

To validate our method, we train a deconvolutional network [17] on PASCAL which takes the image patch inside a given rectangle as input and output the corresponding object mask. The network has the same structure as ours except its input only has RGB channels.

| Methods | Error rate (%) |
|---|---|
| GrabCut [19] | 8.1 |
| BoxPrior [12] | 3.7 |
| OneCut [22] | 6.7 |
| MILCut [24] | 3.6 |
| KernelCut [23] | 7.1 |
| Deconvolution [17] | 4.6 |
| Ours-PASCAL | 4.5 |
| **Ours-COCO** | **3.3** |

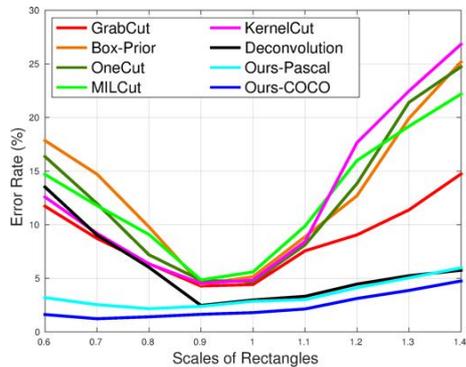

Table 1: Error rates on the provided boxes.　　Figure 4: Error rates on different scales of rectangles.

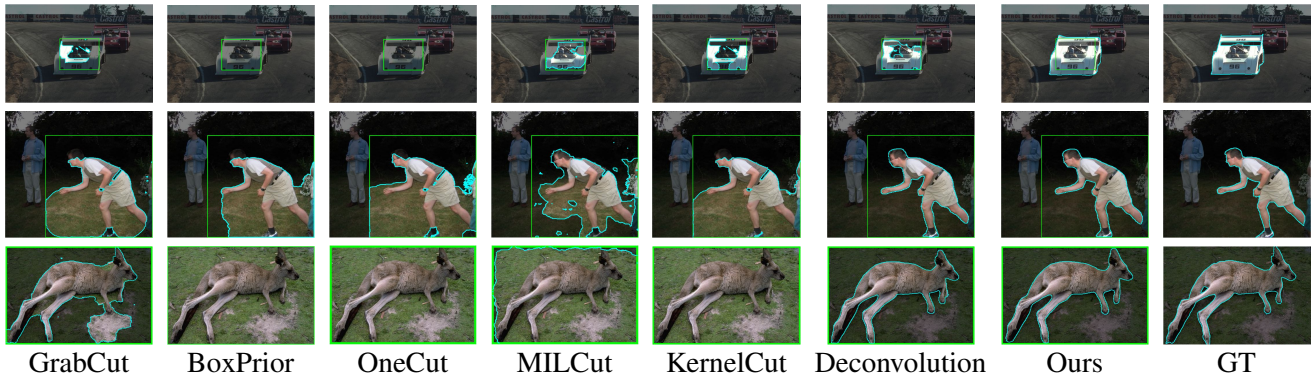

Figure 5: The segmentation results of different methods on the GrabCut dataset.

## 5.2　Rectangle Segmentation

We evaluate both our performance as an interactive selection tool on the GrabCut dataset [19] as well as our performance segmenting objects from detections on the SBD dataset [20]. We report the test results on MS COCO in the supplementary materials [1].

**GrabCut dataset:** For each object of its 50 test images, [12] provides a tight bounding box. The evaluation metric is the error rate which computes the percentage of misclassified pixels within the bounding boxes. We report our results by using these boxes in Table 1. Our PASCAL model is trained on 20 categories, but still has a good generalization ability on unseen objects of this dataset. Our COCO model is trained on more categories and images and thus achieves better results.

Besides using the provided bounding boxes, for each box, we fix its center position and change its size by different scales. We then evaluate all the methods on the generated rectangles. The evaluation metric is the percentage of misclassified pixels over the whole image region. All the baseline implementations are based on the authors' provided codes. The results are illustrated in Fig. 4. It is obvious that previous methods deteriorate their performance rapidly when the rectangle sizes change. Because these methods require the rectangle to bound the object, so either smaller (Fig. 5, rows 1) or larger (Fig. 5, row 2) rectangles cause the method to fail. In the extreme case, when a bounding box covers everything in an image, most methods will select everything as foreground (Fig. 5, row 3). The proposed deconvolutional baseline only performs well when the rectangles are larger because it too requires the rectangle to bound the object. In contrast, our method has consistently good performance with varying rectangle sizes.

**PASCAL dataset:** We run our PASCAL model and previous methods (with code publicly available) on the same detection results obtained by [5] which has a similar detection

---

[1]Project website: https://sites.google.com/view/deepgrabcut

| Methods | $mAP^{r=0.5}$ | $mAP^{r=0.7}$ | $mAP^r_{vol}$ |
|---|---|---|---|
| SDS [8] | 49.7 | 25.3 | 41.4 |
| CFM [4] | 60.7 | 39.6 | - |
| MPA [16] | 61.8 | - | 52.0 |
| IIS [13] | 63.6 | 43.3 | - |
| R2-IOS [14] | **68.8** | 47.5 | - |
| MNC [5] | 65.0 | 46.3 | 55.6 |
| GrabCut [19] | 37.2 | 18.3 | 32.2 |
| Hypercolumn [9] | 62.2 | 41.8 | 52.0 |
| Deconvolution [17] | 59.3 | 39.0 | 50.4 |
| Deconvolution [17] (*enlarged boxes*) | 62.3 | 45.4 | 54.1 |
| **Ours-PASCAL** | 67.3 | **51.0** | **58.6** |

Table 2: Results on the PASCAL validation dataset.

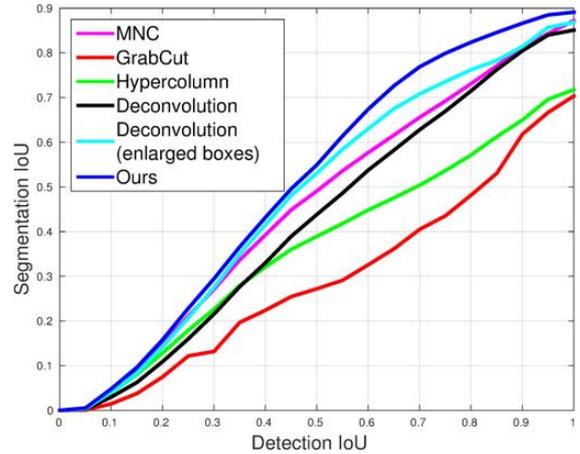

Figure 6: Segmentation IOU *v.s.* Detection IOU.

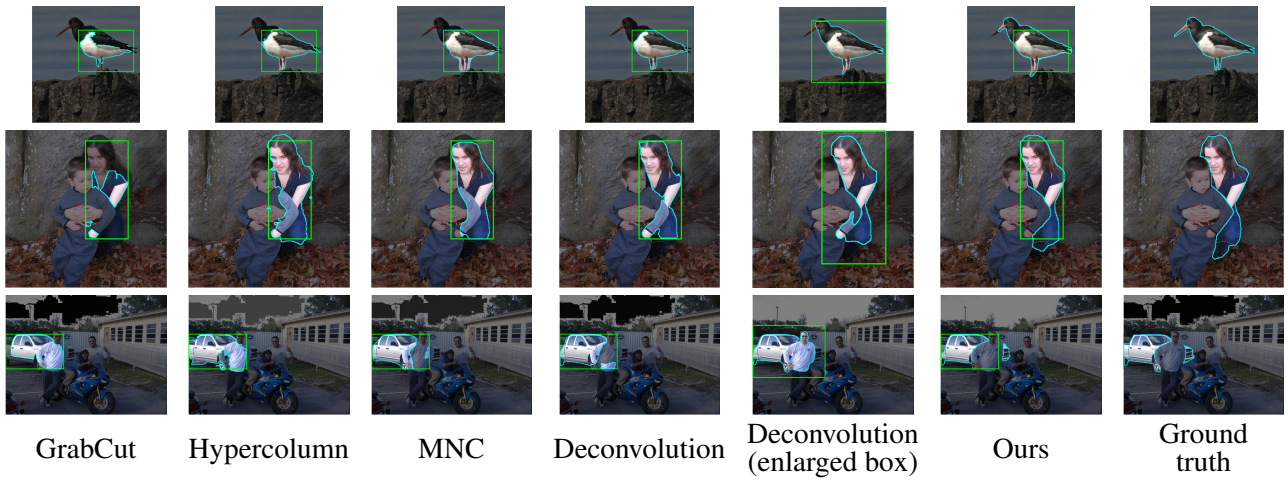

GrabCut    Hypercolumn    MNC    Deconvolution    Deconvolution (enlarged box)    Ours    Ground truth

Figure 7: Segmentation given the same detections on the PASCAL validation set.

framework as Faster R-CNN [18]. All the methods are evaluated by the standard metrics $mAP^r$ at 0.5 and 0.7 thresholds and the $mAP^r_{vol}$ metric which average the $mAP^r$ over 0.1 to 0.9 thresholds. The results are displayed in Table 2. The five methods in the top section of Table 2 use their own detections since we could not get their codes. The methods in the second section use the detection results from [5]. For the proposed deconvolutional baseline, we also compare its performance with enlarged detection boxes (*i.e.* each box is upscaled by 1.5 ratio) to include more context.

It can be seen that given the same detections, our method has the best results under all the evaluation metrics. In overall comparison, our method still achieves the best results under two metrics, which are more related to the segmentation performance. It is also worthy noting that some of the methods [9, 13] can only segment the 20 categories of the PASCAL dataset while our model is generalizable to all object classes. In addition, the deconvolutional baseline has much worse performance than ours, which well demonstrates the importance of our distance-transformation channel.

In Fig. 7 we show some visual results which can intuitively explain why our method works better. In the first row, the detection rectangle only covers parts of the object. All the other comparison methods use the rectangle as hard constraints, and thus fail to segment the whole objects. In the 2nd and 3rd rows, the objects inside the detection boxes have some overlap with other objects. Therefore only using image context inside the rectangles will lose important information. For example, the arm of the boy looks like the arm of the woman inside the box (2nd row) and the separate part of the car is impossible to be inferred without having the global view (3rd row). Therefore our novel way of leveraging rectangle information is the key for the success of our method.

| metric | Segmentation models | $v=0$ | 0.1 | 0.2 | 0.3 |
|---|---|---|---|---|---|
| $mAP^{r=0.5}$ | GrabCut [19] | 34.2 | 30.5 | 23.1 | 13.6 |
| | Hypercolumn [9] | 60.0 | 58.3 | 55.1 | 43.5 |
| | **Ours** | **64.0** | **63.7** | **62.1** | **57.3** |
| $mAP^{r=0.7}$ | GrabCut [19] | 16.7 | 12.3 | 5.9 | 2.5 |
| | Hypercolumn [9] | 40.4 | 34.5 | 25.9 | 13.1 |
| | **Ours** | **45.5** | **44.8** | **43.1** | **36.4** |
| $mAP^r_{vol}$ | GrabCut [19] | 30.9 | 28.2 | 23.3 | 17.7 |
| | Hypercolumn [9] | 51.0 | 49.2 | 45.6 | 40.0 |
| | **Ours** | **55.0** | **54.8** | **53.7** | **49.8** |

Table 3: Evaluation of rectangle misplacement.

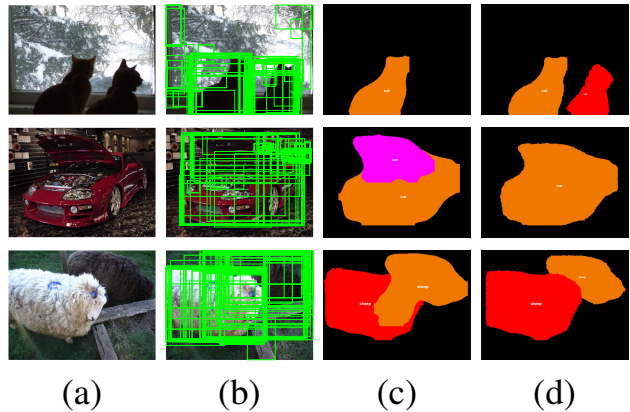

(a) (b) (c) (d)

Figure 8: (a) Input image. (b) All detections of [5]. (c) Using the naive labeling approach (score thresholding at 0.7). (d) Using our labeling approach.

To further evaluate our method, we first compare the intersection-over-union (IOU) score of the detection rectangles of [5] with the IOU of the corresponding segmentation. Each detection/segmentation is compared to the ground truth rectangle/segment that has the highest detection IOU. Then all the rectangle IOU scores and mask IOU scores are averaged over 0.05 interval bins. The results are illustrated in Fig. 6. The curve from our method is above all the other methods, indicating that our method computes the best segmentations given the same rectangles. It is also worth noting that our method has a higher IOU for its segmentations than its detections (*e.g.* when the rectangle IOU $\approx 0.6$ our mask IOU $\approx 0.7$).

In the second experiment in Table 3, we randomly jitter the detections of [9] using Eq. 1 with the uniform distribution $[-1,1]$ and evaluate the results under the standard metrics. Our results show much less degradation with the increasing parameter *v* compared to two other methods, demonstrating that our method is more robust to the rectangle placement.

## 5.3 Instance-level semantic segmentation

We evaluate our pixel labeling approach on the PASCAL validation dataset given a set of detections. We compare to a naive baseline used in [5] for visualizing their segmentation results that thresholds segments by their detection scores and then paints each segment into the final label map. Since our labeling method is generic to all methods which can output segmentation probability maps, we first compare the results by using the segmentations of [5] in Fig. 8.

There are some obvious limitations for the naive labeling approach: 1) objects with small detection scores will be eliminated (Fig. 8, row 1), 2) it will get confused when an object is covered by multiple detections (Fig. 8, row 2). 3) It does not leverage the information of labeling consistency between nearby pixels (Fig. 8, row3). While our labeling approach can solve all the problems. More visual results obtained by our rectangle segmentations and our labeling approach are shown in Fig. 9.

## 5.4 Handling arbitrary closed curves

We have demonstrated the effectiveness of our segmentation model on sloppy rectangle inputs. In fact, our model can even generalize to arbitrary closed curves (such as circle, ellipse,

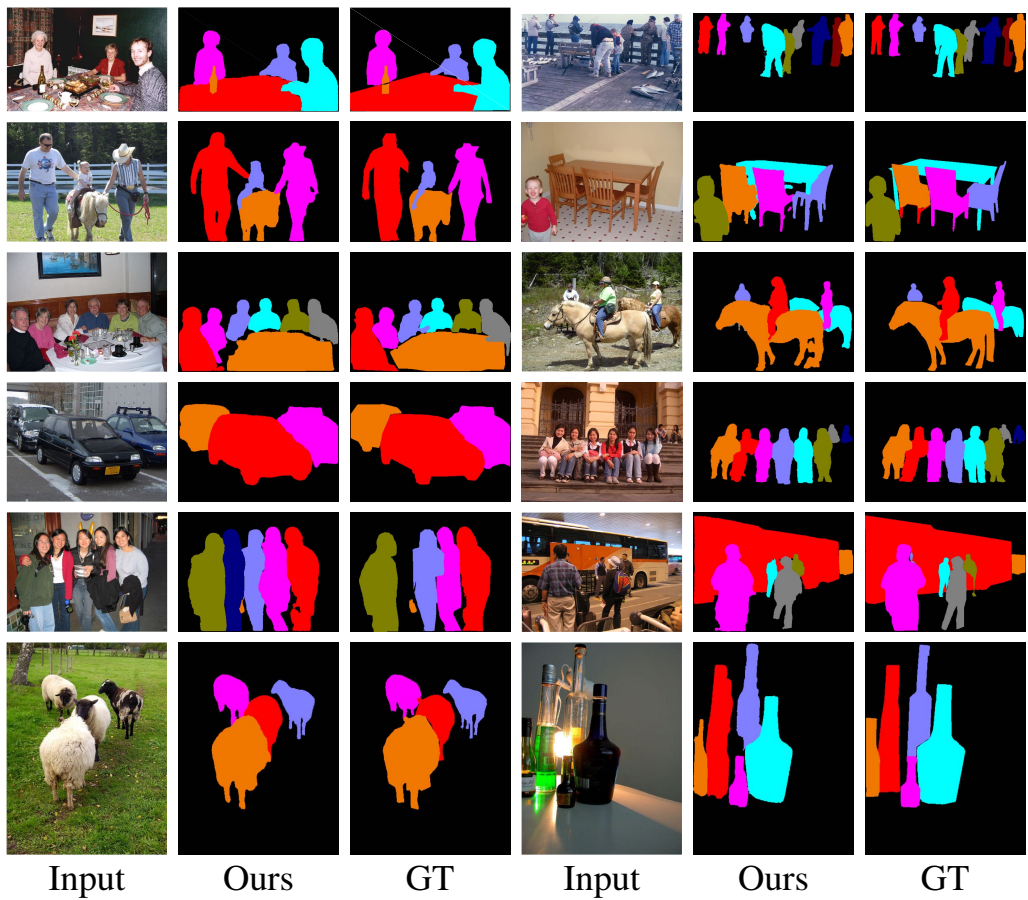

Input    Ours    GT    Input    Ours    GT

Figure 9: The visual results of our pixel-level labeling based on our rectangle segmentations.

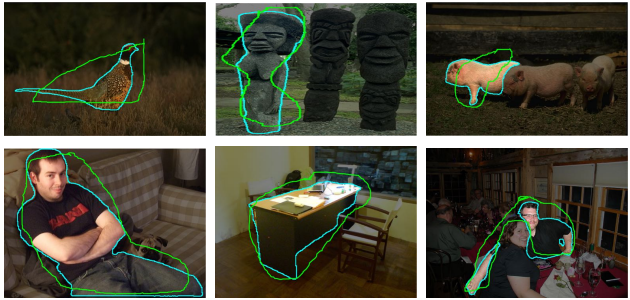

Figure 10: In each image, the user selection is outlined in green and the segmentation is outlined in cyan.

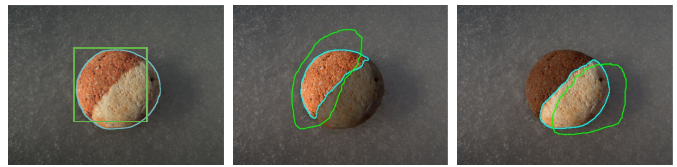

Figure 11: While a bounding box cannot select one half of the stone, loosely-drawn curves easily can.

triangles *etc*.), even though our model is only trained on rectangles. Following a similar testing procedure for a rectangle input, an arbitrary closed curve (generated either manually or automatically) is transformed into a distance map as pixels along the edges of the curve have the distance 128 and the distances of the other pixels are computed as Eqn. 2. We show several visual examples in Fig. 10 as well as some video demos in the supplementary materials to prove the great flexibility of our approach.

Allowing arbitrary closed shapes not only allows greater flexibility and arguably a more natural user interface, but also can provide more accurate results, especially in the case where multiple objects share similar bounding boxes. In Fig. 11 a bounding box is placed around the darker half of a stone in an attempt to select just that half. However, this bounding box almost entirely includes the lighter half, so our method selects the entire stone. However, a loosely-drawn curve around either half of the stone can accurately select the respective half. Our model extends to allow such flexibility without requiring retraining.

## 6 Conclusions

We propose a novel neural network based segmentation method that uses rectangles as input. Our segmentation model is robust to the placement of the rectangles and has a good understanding of global context. Moreover, our method generalizes well to curve-based inputs without retaining. We also propose a CRF-based labeling approach for the instance-level semantic segmentation task to have a unique labeling of each pixel. Experimental results on several benchmark datasets demonstrate the effectiveness of the proposed method on both the interactive segmentation and instance segmentation tasks.